\documentclass[runningheads]{llncs}
\usepackage[T1]{fontenc}
\usepackage{multirow}
\usepackage{graphicx,verbatim}
\usepackage{amsmath}
\usepackage{amsfonts}
\usepackage{subcaption}
\usepackage{footmisc}
\begin{document}

\title{UltrasoundAgents: Hierarchical Multi-Agent Evidence-Chain Reasoning for Breast Ultrasound Diagnosis}
\titlerunning{UltrasoundAgents}
%

\authorrunning{Zhu et al.}

\author{Yali Zhu\inst{1,2}\thanks{Contributed equally.} \and
Kang Zhou\inst{3}$^*$ \and
Dingbang Wu\inst{1,2} \and
Gaofeng Meng\inst{1,2,3}}
%

%
\institute{
Institute of Automation, Chinese Academy of Sciences, Beijing, China \and
School of Advanced Interdisciplinary Sciences, University of Chinese Academy of Sciences, Beijing, China \and
Centre for Artificial Intelligence and Robotics, Hong Kong Institute of Science \& Innovation, Chinese Academy of Sciences, Hong Kong \\
\email{\{kang.zhou, gaofeng.meng\}@cair-cas.org.hk}
}




\maketitle              

\begin{abstract}
Breast ultrasound diagnosis typically proceeds from global lesion localization to local sign assessment and then evidence integration to assign a BI-RADS category and determine benignity or malignancy. Many existing methods rely on end-to-end prediction or provide only weakly grounded evidence, which can miss fine-grained lesion cues and limit auditability and clinical review. To align with the clinical workflow and improve evidence traceability, we propose a hierarchical multi-agent framework, termed UltrasoundAgents. A main agent localizes the lesion in the full image and triggers a crop-and-zoom operation. A sub-agent analyzes the local view and predicts four clinically relevant attributes, namely echogenicity pattern, calcification, boundary type, and edge (margin) morphology. The main agent then integrates these structured attributes to perform evidence-based reasoning and output the BI-RADS category and the malignancy prediction, while producing reviewable intermediate evidence. Furthermore, hierarchical multi-agent training often suffers from error propagation, difficult credit assignment, and sparse rewards. To alleviate this and improve training stability, we introduce a decoupled progressive training strategy. We first train the attribute agent, then train the main agent with oracle attributes to learn robust attribute-based reasoning, and finally apply corrective trajectory self-distillation with spatial supervision to build high-quality trajectories for supervised fine-tuning, yielding a deployable end-to-end policy. Experiments show consistent gains over strong vision-language baselines in diagnostic accuracy and attribute agreement, together with structured evidence and traceable reasoning.

\end{abstract}

\section{Introduction}
Breast ultrasound (BUS) is widely used for breast cancer screening and early diagnosis. Clinicians typically follow a coarse-to-fine reading workflow: localize a suspicious lesion, examine fine-grained signs (e.g., morphology and echo patterns), and integrate evidence to assign a BI-RADS category and decide benignity or malignancy \cite{biradsnet,buscot,spak2017birads}. A practical computer-aided diagnosis system therefore needs not only accurate predictions, but also intermediate evidence that is traceable and consistent with the clinical decision process \cite{interpretable}.

Most BUS methods use end-to-end prediction or tightly coupled multi-stage pipelines \cite{dualstage}. While interpretable multi-task and multi-stage designs have been explored \cite{biradsnet,interpretable}, the produced evidence is often only weakly explicit, lacking a clear evidence-chain to explain how local cues support BI-RADS grading and malignancy decisions.
Predicting BI-RADS categories and/or generating BI-RADS descriptor terms as structured outputs has also been studied \cite{hayashida2022birads4a,carrilero2024describe}, but these approaches are usually trained with static supervision and do not explicitly model the dynamic, sequential evidence acquisition and integration process inherent in clinical practice.

Recent vision-language models (VLMs) enable more explicit reasoning via textual rationales and structured attribute descriptions \cite{buscot}. However, supervised fine-tuning (SFT) for reasoning often depends on additional rationale annotations or teacher trajectories and may inherit biases from static supervision. Reinforcement learning (RL) can instead optimize towards outcome-level rewards and encourage the model to discover readable reasoning patterns without reference rationales \cite{medr1,medvlmr1}. For medical images, where key evidence can be small and sparse, incorporating ROI-focused observations during reasoning is also effective, and RL can incentivize such behavior in an end-to-end manner \cite{medreason,deepeyes,vitar}. Nevertheless, learning localization, fine-grained perception, and high-level diagnostic reasoning within a single policy remains difficult: localization errors and perceptual noise shift downstream observations, increasing non-stationarity and complicating credit assignment \cite{pateria2021hrlsurvey}. Hierarchical multi-agent diagnosis provides a principled alternative by separating evidence acquisition from evidence integration, improving consistency and traceability \cite{medagent,treereasoning}, yet stable training and distillation into deployable policies remain underexplored for unified BUS diagnosis.

We propose \textbf{UltrasoundAgents}, a hierarchical multi-agent BUS diagnostic framework aligned with the clinical workflow. We explicitly decouple global localization and high-level evidence integration from fine-grained attribute perception. The main agent localizes lesions on the full image and integrates evidence to output BI-RADS and malignancy conclusions. The sub-agent focuses on the crop-and-zoom lesion view, recognizes clinically relevant attributes, and provides structured evidence to support the main agent. To ensure stability and robustness of hierarchical training, we further adopt a decoupled progressive training scheme. We first train an attribute specialist, then introduce oracle-guided curriculum reinforcement learning for the main agent to mitigate non-stationarity in hierarchical training, and finally construct high-quality supervised trajectories via corrective trajectory self-distillation and distill them into a deployable policy through SFT. At test time, the system relies only on predicted intermediate evidence while maintaining a consistent evidence chain and robust diagnostic decisions.

\begin{figure}[t]
    \centering
    \includegraphics[width=0.95\linewidth]{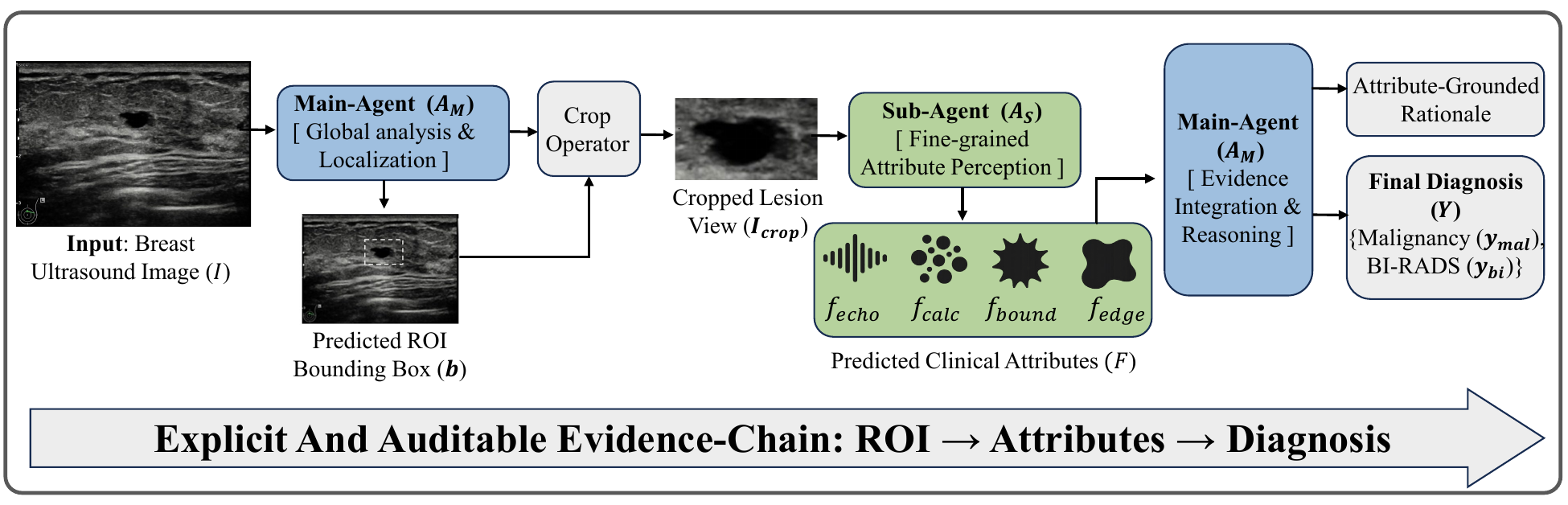}
    \caption{Hierarchical multi-agent architecture. The main agent analyzes the full image to localize the lesion, triggers a crop-and-zoom operation, and queries the sub-agent on the zoomed view to obtain structured attribute evidence. The main agent then integrates the global context and the attributes to predict malignancy and BI-RADS.
    }
    \label{fig:overview}
\end{figure}

This work makes three contributions. (1) We propose a hierarchical multi-agent diagnostic framework that separates full-image lesion localization and high-level evidence-based reasoning in a main agent from local clinical attribute recognition in a sub-agent, using crop-and-zoom to produce an auditable chain of ROI $\rightarrow$ attributes $\rightarrow$ diagnosis. To our knowledge, we present the first agent-based diagnostic framework for ultrasound diagnosis. (2) We introduce an oracle-guided curriculum RL strategy that trains the main agent with ground-truth attributes to learn stable attribute-based reasoning without perceptual noise, alleviating non-stationarity and error propagation. (3) We design a corrective trajectory self-distillation pipeline that refines RL-explored trajectories into supervised data and distills them via SFT into a deployable policy, improving reasoning consistency and localization-aware perception when predicted attributes are used at test time.

\section{Method}
\label{sec:method}

\begin{figure}[t]
    \centering
    \includegraphics[width=0.95\linewidth]{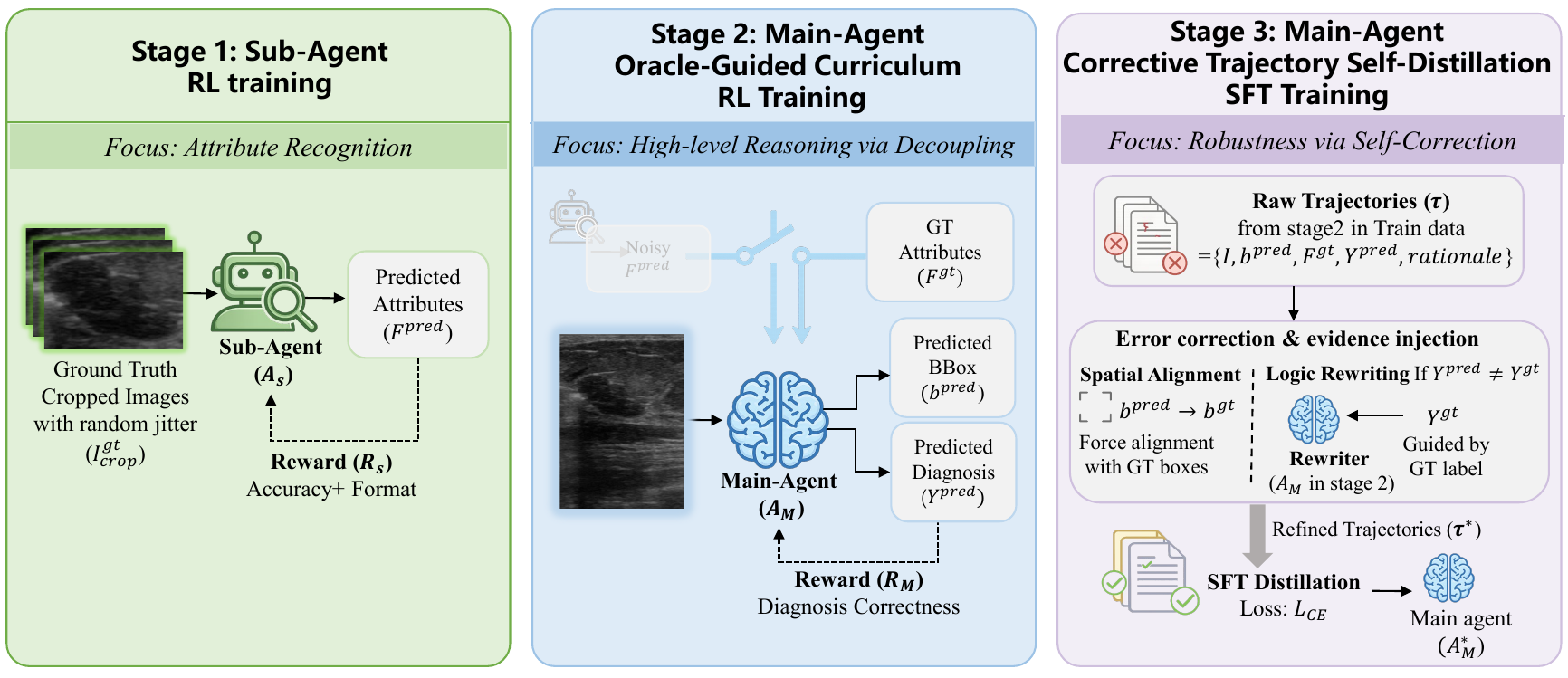}
    \caption{Three-stage training. Stage 1: RL trains $A_S$ for attribute recognition. Stage 2: oracle-guided RL trains $A_M$ with GT attributes for stable reasoning. Stage 3: refine trajectories and distill via SFT for robustness.}
    \label{fig:train}
\end{figure}

As shown in Fig.~\ref{fig:overview}, we propose a hierarchical multi-agent framework for breast ultrasound interpretation that follows a coarse-to-fine clinical workflow. The main design decouples global localization and high-level reasoning from fine-grained attribute perception, and uses an explicit crop-and-zoom operation to provide focused lesion observations. The framework contains a main agent $A_M$ for full-image analysis, lesion localization, and evidence integration, and a sub-agent $A_S$ specialized in attribute recognition. Training proceeds in three stages: RL for $A_S$, oracle-guided curriculum RL for $A_M$, and trajectory refinement followed by supervised fine-tuning (SFT), as illustrated in Fig.~\ref{fig:train}.

\subsection{Hierarchical multi-agent architecture}
\label{sec:method:arch}

Fig.~\ref{fig:overview} summarizes the interaction between the two agents. The input is a breast ultrasound image $I \in \mathbb{R}^{H \times W \times C}$. The outputs include a malignancy label $y_{\text{mal}}$ and a BI-RADS category $y_{\text{bi}}$. Each image is associated with a ground-truth lesion box $b^{\text{gt}}$ and four attributes: echogenicity pattern $f_{echo}$, calcification $f_{calc}$, boundary type $f_{bound}$, and edge (margin) morphology $f_{edge}$.

\noindent
\textbf{Main agent ($A_M$): global localization and reasoning}
Given $I$, $A_M$ performs global scanning and predicts an ROI box $b=[x_1,y_1,x_2,y_2]$ , then triggers crop-and-zoom to obtain a local lesion view
\begin{equation}
I_{\text{crop}} = \mathrm{Crop}(I, b).
\end{equation}
After receiving the attribute set from $A_S$, $A_M$ combines global context from $I$ with local clinical attributes and outputs the diagnosis $Y=\{y_{\text{mal}}, y_{\text{bi}}\}$. 

\noindent
\textbf{Sub-agent ($A_S$): fine-grained attribute expert}
$A_S$ operates on $I_{\text{crop}}$ and predicts
\begin{equation}
F=\{f_{echo}, f_{calc}, f_{bound}, f_{edge}\}.
\end{equation}
We adopt the attribute taxonomy from BUS-CoT \cite{buscot}. These attributes
serve as structured clinical evidence provided to $A_M$. This explicit division of labor reduces the learning burden of $A_M$ and improves interpretability.

\subsection{Decoupled training and trajectory self-distillation}
\label{sec:method:stage1}

\noindent
\textbf{Stage 1: RL for the sub-agent.}
We train $A_S$ to produce reliable attributes from $I_{\text{crop}}$ with readable reasoning traces. We use reinforcement learning with a reward that encourages both attribute correctness and compliance with a predefined template, 
\begin{equation}
R_S = \mathrm{Acc}(F^{\text{pred}}, F^{\text{gt}}) + R_{\text{fmt}},
\end{equation}
where $\mathrm{Acc}(\cdot)$ measures the agreement with the annotations and $R_{\text{fmt}}$ scores formatting validity. This stage provides an attribute specialist that is easier to train and yields interpretable intermediate evidence.

\noindent
\textbf{Stage 2: oracle-guided curriculum RL for the main agent.}
Training $A_M$ end-to-end with imperfect attributes predicted by the sub-agent is often unstable. Early attribute noise makes the observation distribution of the main agent non-stationary and propagates errors through the hierarchy. To decouple reasoning learning from perception errors, we introduce an oracle-guided curriculum RL for $A_M$. Specifically, during this stage we replace the sub-agent outputs with ground-truth intermediate evidence, so that $A_M$ learns attribute-based diagnostic reasoning under a relatively clean intermediate state. The reward for the main agent focuses only on diagnostic correctness,
\begin{equation}
R_M = \lambda_1 \cdot \mathbb{I}[y_{\text{mal}}^{\text{pred}}=y_{\text{mal}}^{\text{gt}}] + \lambda_2 \cdot \mathbb{I}[y_{\text{bi}}^{\text{pred}}=y_{\text{bi}}^{\text{gt}}],
\end{equation}
where $\lambda_1,\lambda_2$ weight malignancy and BI-RADS.

This stage does not rely on sparse diagnostic reward to directly learn precise localization. Localization is mainly improved in Stage 3. Oracle evidence is used only in training; at test time $A_M$ consumes predicted attributes from $A_S$.

\noindent
\textbf{Stage 3: corrective self-distillation via trajectory refinement.}
Since Stage 2 provides limited supervision for localization and RL rollouts are noisy, we distill the learned behavior into a deployable main agent using trajectory correction and refinement followed by SFT. We sample trajectories from the Stage 2 policy,
\[
\tau=\{(I, b^{\text{pred}}, F^{\text{gt}}, Y^{\text{pred}}, \text{rationale}^{\text{pred}})\}.
\]
We then refine them by injecting reliable supervision. For all samples, we replace the predicted box with the ground-truth box, $b^{\text{pred}}\leftarrow b^{\text{gt}}$, to enforce spatial grounding. For samples with incorrect diagnosis ($Y^{\text{pred}}\neq Y^{\text{gt}}$), we use the Stage 2 model as a rewriter to regenerate the rationale conditioned on the ground-truth label $Y^{\text{gt}}$. This yields refined trajectories $\tau^*$, which form a supervised set $\mathcal{D}^*=\{\tau^*\}$ for cross-entropy SFT. This process turns sparse and noisy RL rollouts into denser supervision signals, strengthens localization, and transfers the stable attribute-based reasoning learned in Stage 2 into a deployable policy.
\[
\tau^*=\mathrm{Refine}\big(\tau \mid b \leftarrow b^{\text{gt}},\; \text{if } Y^{\text{pred}}\neq Y^{\text{gt}}\text{ then rewrite rationale with }Y^{\text{gt}}\big).
\]

\section{Experiments}
\label{sec:exp}

\subsection{Setups}
\label{sec:exp:data}

\textbf{Datasets.}
We evaluate in-domain performance on three public breast ultrasound datasets, BUSBRA \cite{busbra}, BUSI \cite{busi}, and BUDIAT \cite{budiat}, and test out-of-domain (OOD) generalization on BrEaST \cite{breast}. Since fine-grained attribute annotations are limited in public BUS data, we follow the unified taxonomy and split protocol released by BUS-CoT, which provides lesion bounding boxes, malignancy labels, BI-RADS categories, and fine-grained ultrasound attributes for these sources \cite{buscot}. BrEaST is an independent OOD dataset with attribute annotations, and we construct consistent attribute labels on BrEaST under the same BUS-CoT taxonomy for OOD evaluation \cite{buscot}.

\noindent
\textbf{Task definition.}
Given a breast ultrasound image, the model predicts malignancy and BI-RADS category. We also evaluate the intermediate evidence-chain by reporting IoU between predicted and ground-truth lesion boxes and recognition performance on four clinically relevant attributes.

\noindent
\textbf{Evaluation metrics.}
For malignancy prediction, we report ROC-AUC (AUC) and accuracy (Acc). For BI-RADS grading, we report BI-RADS accuracy (Bi-Acc) and Cohen's $\kappa$ \cite{kappa}. Localization is evaluated by IoU. For each attribute task, we report accuracy, Macro-F1, and Weighted-F1.

\noindent
\textbf{Implementation details.}
\label{sec:exp:impl}
All methods use the same base VLM, Qwen2.5-VL-3B (abbrev. Qwen2.5-3B). Images keep the original aspect ratio and are resized to at most $600\times 800$, while the crop-and-zoom view is no smaller than $224\times224$. We follow the three-stage training procedure in Sec.~\ref{sec:method}. Both RL stages use GRPO \cite{shao2024deepseekmath} with temperature 0.8, learning rate $1\times10^{-6}$, batch size 64, and 8 samples per prompt on two NVIDIA L40S GPUs. The SFT stage uses a learning rate of $1\times10^{-5}$ with \text{per\_device\_batch\_size}=4.

\noindent
\textbf{Baselines.}
\label{sec:exp:baseline}
All baselines use the same multimodal backbone (Qwen2.5-3B) and differ only in training strategy and inference workflow. \textit{Zero-shot} directly prompts the model on the full image to predict malignancy and BI-RADS. \textit{CoT-SFT }follows BUS-CoT \cite{buscot} and uses SFT on template-based reasoning data with intermediate attribute evidence. \textit{Think-with-Image} follows the DeepEyes-style single-agent workflow \cite{deepeyes}, which localizes an ROI on the full image, applies crop-and-zoom, and then re-analyzes the cropped view for attributes and diagnosis. 

\subsection{Evaluation results}
\label{sec:exp:main}

\textbf{Overall performance.}
Table~\ref{tab:main} reports overall diagnostic performance on three in-domain subsets and one OOD test set.

\begin{table}[t]
\centering
\caption{Main diagnostic performance on four datasets. Overall is computed on the pooled set across all datasets (i.e., sample-wise micro average).}
\label{tab:main}
\resizebox{\textwidth}{!}{%
\begin{tabular}{l|cccc|cccc|cccc|cccc|cccc}
\hline
\multirow{2}{*}{Method} & \multicolumn{4}{c|}{BUSBRA} & \multicolumn{4}{c|}{BUSI} & \multicolumn{4}{c|}{BUDIAT} & \multicolumn{4}{c|}{BrEaST (OOD)} & \multicolumn{4}{c}{Overall} \\
& AUC & Acc & Bi-Acc & $\kappa$ & AUC & Acc & Bi-Acc & $\kappa$ & AUC & Acc & Bi-Acc & $\kappa$ & AUC & Acc & Bi-Acc & $\kappa$ & AUC & Acc & Bi-Acc & $\kappa$ \\
\hline
Qwen2.5-3B-Zero-Shot & 0.458 & 0.588 & 0.091 & 0.003 & 0.493 & 0.573 & 0.156 & 0.066 & 0.500 & 0.743 & 0.200 & -0.053 & 0.484 & 0.608 & 0.078 & -0.065 & 0.476 & 0.602 & 0.117 & 0.014 \\
Qwen2.5-3B-COT-SFT & 0.668 & 0.722 & 0.563 & 0.258 & \textbf{0.835} & \textbf{0.833} & 0.46 & \textbf{0.245} & 0.722 & 0.857 & \textbf{0.4} & 0.129 & 0.586 & 0.627 & 0.176 & \textbf{0.06} & 0.71 & 0.751 & 0.468 & 0.204 \\
Think-with-Image & 0.500 & 0.693 & 0.048 & 0.003 & 0.528 & 0.660 & 0.080 & 0.019 & 0.500 & 0.743 & 0.314 & -0.001 & 0.526 & 0.647 & \textbf{0.196} & 0.005 & 0.512 & 0.683 & 0.101 & 0.004 \\
ours & \textbf{0.723} & \textbf{0.813} & \textbf{0.62} & \textbf{0.3} & 0.784 & \textbf{0.833} & \textbf{0.542} & 0.244 & \textbf{0.778} & \textbf{0.886} & \textbf{0.4} & \textbf{0.145} & \textbf{0.685} & \textbf{0.725} & 0.157 & 0.037 & \textbf{0.741} & \textbf{0.813} & \textbf{0.515} & \textbf{0.224} \\
\hline
\end{tabular}
}
\end{table}

\textit{In-domain results.}
Overall, Our method achieves the best performance (AUC 0.741, Acc 0.813, Bi-Acc 0.515, $\kappa$ 0.224), outperforming the strongest baseline CoT-SFT. The crop-and-zoom operation alone does not guarantee improvement: although Think-with-Image includes a crop-then-revisit workflow, its performance is close to random in Table~\ref{tab:main}. In contrast, our explicit division of labor in the ROI$\rightarrow$attribute$\rightarrow$diagnosis evidence-chain, together with oracle-guided curriculum RL training and corrective trajectory self-distillation, allows the cropped local view to be reliably converted into structured attribute evidence and consistently used by the main agent for joint BI-RADS and malignancy decisions. Across BUSBRA, BUSI, and BUDIAT, Ours remains competitive or better on both diagnostic tasks, indicating stable in-domain behavior.

\textit{Out-of-domain generalization.}
On BrEaST (OOD), our method improves malignancy diagnosis over CoT-SFT (AUC 0.685 vs 0.586, Acc 0.725 vs 0.627). Tab.~\ref{tab:feature_acc_f1} shows that lesion crops yield higher F1 scores than whole-image inputs for Boundary, Edge, and Echo on BrEaST, which indicates more reliable fine-grained attribute evidence under imaging-style shift. We attribute the OOD gain to this ROI-focused crop-and-zoom workflow, which reduces background sensitivity and alleviates the effective resolution limit of full-image inputs. BI-RADS grading remains more challenging across domains, and the GTbox and GTattr upper bounds in Tab.~\ref{tab:ablation} suggest that localization bias and attribute noise are the main bottlenecks for OOD consistency.

\noindent
\textbf{Ablation and upper-bound analysis.}
\label{sec:exp:ablation}
Table~\ref{tab:ablation} reports ablations, upper bounds with ground-truth boxes (GTbox) or attributes (GTattr), and localization IoU.
\begin{table}[t]
\centering
\caption{Ablation study and upper-bound analysis. }
\label{tab:ablation}
\resizebox{\textwidth}{!}{%
\begin{tabular}{l|ccccc|ccccc|ccccc|ccccc|ccccc}
\hline
\multirow{2}{*}{Method} & \multicolumn{5}{c|}{BUSBRA} & \multicolumn{5}{c|}{BUSI} & \multicolumn{5}{c|}{BUDIAT} & \multicolumn{5}{c|}{BrEaST (OOD)} & \multicolumn{5}{c}{Overall} \\
& AUC & Acc & Bi-Acc & $\kappa$ & IoU & AUC & Acc & Bi-Acc & $\kappa$ & IoU & AUC & Acc & Bi-Acc & $\kappa$ & IoU & AUC & Acc & Bi-Acc & $\kappa$ & IoU & AUC & Acc & Bi-Acc & $\kappa$ & IoU \\
\hline
w/o Oracle Training & 0.500 & 0.693 & 0.497 & 0.043 & 0.266 & 0.569 & 0.690 & 0.430 & 0.020 & 0.406 & 0.556 & 0.771 & 0.371 & 0.055 & 0.411 & 0.563 & 0.667 & 0.098 & -0.013 & 0.346 & 0.535 & 0.696 & 0.413 & 0.018 & 0.328 \\
w/o Self-Distill & 0.679 & 0.738 & 0.524 & 0.175 & 0.254 & 0.771 & 0.802 & 0.521 & 0.241 & 0.373 & \textbf{0.889} & \textbf{0.943} & \underline{0.400} & 0.159 & 0.348 & 0.686 & 0.686 & 0.137 & 0.012 & 0.294 & 0.726 & 0.767 & 0.458 & 0.173 & 0.299 \\
ours & 0.723 & 0.813 & \underline{0.620} & \underline{0.300} & \underline{0.638} & \underline{0.784} & \underline{0.833} & \underline{0.542} & \underline{0.244} & \underline{0.610} & \underline{0.778} & \underline{0.886} & \underline{0.400} & 0.145 & \underline{0.567} & 0.685 & 0.725 & 0.157 & 0.037 & \underline{0.537} & 0.741 & 0.813 & \underline{0.515} & \underline{0.224} & \underline{0.610} \\
ours-with-GTbox & \textbf{0.780} & \underline{0.845} & 0.583 & 0.243 & \textbf{1} & 0.769 & 0.823 & \underline{0.542} & 0.229 & \textbf{1} & \underline{0.778} & \underline{0.886} & \textbf{0.429} & \underline{0.176} & \textbf{1} & \textbf{0.801} & \textbf{0.804} & \underline{0.176} & \underline{0.045} & \textbf{1} & \underline{0.782} & \underline{0.837} & 0.501 & 0.208 & \textbf{1} \\
ours-with-GTattr& \underline{0.778} & \textbf{0.856} & \textbf{0.668} & \textbf{0.391} & 0.622 & \textbf{0.851} & \textbf{0.884} & \textbf{0.663} & \textbf{0.459} & 0.560 & 0.759 & 0.857 & \textbf{0.429} & \textbf{0.181} & 0.560 & \underline{0.796} & \underline{0.784} & \textbf{0.216} & \textbf{0.086} & 0.395 & \textbf{0.804} & \textbf{0.853} & \textbf{0.582} & \textbf{0.345} & 0.568 \\
\hline
\end{tabular}
}
\end{table}

\begin{table}[t]
\centering
\caption{Attribute classification accuracy, Macro-F1, and Weighted-F1 for whole vs lesion inputs.Calcification omitted from the table due to severe imbalance.}
\label{tab:feature_acc_f1}
\resizebox{\textwidth}{!}{%
\begin{tabular}{l|ccc|ccc|ccc|ccc|ccc}
\hline
\multirow{2}{*}{Attribute (Input)} & \multicolumn{3}{c|}{BUSBRA} & \multicolumn{3}{c|}{BUSI} & \multicolumn{3}{c|}{BUDIAT} & \multicolumn{3}{c|}{BrEaST} & \multicolumn{3}{c}{Overall} \\
& Acc & Macro-F1 & W-F1 & Acc & Macro-F1 & W-F1 & Acc & Macro-F1 & W-F1 & Acc & Macro-F1 & W-F1 & Acc & Macro-F1 & W-F1 \\
\hline
Boundary (cotsft) & 0.511 & 0.321 & 0.399 & 0.620 & 0.394 & 0.535 & 0.514 & 0.330 & 0.375 & 0.490 & 0.374 & 0.428 & 0.537 & 0.341 & 0.436 \\
Boundary (whole) & 0.535 & 0.326 & 0.405 & \textbf{0.656} & 0.394 & 0.548 & 0.514 & 0.341 & 0.377 & 0.529 & 0.387 & 0.429 & 0.564 & 0.340 & 0.440 \\
Boundary (lesion) & \textbf{0.567} & \textbf{0.447} & \textbf{0.510} & 0.635 & \textbf{0.419} & \textbf{0.568} & \textbf{0.571} & \textbf{0.472} & \textbf{0.460} & \textbf{0.588} & \textbf{0.524} & \textbf{0.566} & \textbf{0.588} & \textbf{0.445} & \textbf{0.526} \\
\hline
Edge (cotsft) & 0.595 & 0.462 & 0.517 & 0.670 & 0.540 & 0.607 & 0.514 & 0.440 & 0.375 & 0.608 & 0.618 & 0.605 & 0.609 & 0.473 & 0.541 \\
Edge (whole) & 0.615 & 0.452 & 0.512 & 0.708 & 0.541 & 0.618 & 0.514 & 0.455 & 0.377 & 0.667 & 0.662 & 0.655 & 0.637 & 0.475 & 0.546 \\
Edge (lesion) & \textbf{0.647} & \textbf{0.573} & \textbf{0.603} & \textbf{0.781} & \textbf{0.718} & \textbf{0.758} & \textbf{0.571} & \textbf{0.560} & \textbf{0.466} & \textbf{0.726} & \textbf{0.754} & \textbf{0.759} & \textbf{0.686} & \textbf{0.608} & \textbf{0.650} \\
\hline
Echo (cotsft) & 0.587 & 0.153 & 0.523 & 0.430 & 0.300 & 0.386 & 0.657 & 0.264 & 0.703 & 0.569 & 0.259 & 0.496 & 0.549 & 0.168 & 0.487 \\
Echo (whole) & 0.634 & 0.129 & 0.492 & 0.292 & 0.090 & 0.132 & \textbf{0.886} & \textbf{0.313} & \textbf{0.832} & \textbf{0.795} & 0.295 & 0.704 & 0.584 & 0.105 & 0.431 \\
Echo (lesion)  & \textbf{0.656} & \textbf{0.232} & \textbf{0.609} & \textbf{0.563} & \textbf{0.440} & \textbf{0.535} & 0.657 & 0.269 & 0.715 & \textbf{0.795} & \textbf{0.464} & \textbf{0.773} & \textbf{0.646} & \textbf{0.297} & \textbf{0.606} \\
\hline
\end{tabular}
}
\end{table}
\textit{Effect of oracle-guided curriculum RL training.}
Removing oracle training severely degrades diagnostic performance and spatial alignment on the Overall split (AUC 0.741 to 0.535, $\kappa$ 0.224 to 0.018, IoU 0.610 to 0.328), which supports the need to decouple reasoning learning from early-stage attribute noise in a hierarchical system.

\textit{Contribution of Corrective trajectory self-distillation.}
Removing self-distillation yields an Overall IoU of only 0.299, while the final model reaches 0.610. This indicates that corrective trajectory refinement, together with injected ground-truth box supervision, strengthens spatial attention alignment and improves reasoning consistency. Diagnostic performance also improves from Overall AUC 0.726 to 0.741, which suggests that better localization translates into more reliable downstream decisions.

\textit{Error source decomposition via GTbox and GTattr upper bounds.}
Using GTbox improves the overall AUC to 0.782, which highlights localization error as a major factor. Using GTattr further improves overall AUC to 0.804 and $\kappa$ to 0.345, suggesting that attribute noise is a primary bottleneck for BI-RADS consistency.This analysis identifies current bottlenecks and suggests substantial headroom by improving localization and attribute reliability while keeping the same reasoning structure.

\noindent
\textbf{Attribute recognition and the effect of crop-and-zoom.}
\label{sec:exp:attr}
Table~\ref{tab:feature_acc_f1} compares attribute recognition with different inputs (whole image vs.\ lesion crop), using CoT-SFT attribute predictions as a reference. Overall, lesion crops yield consistently higher F1 scores for Boundary, Edge, and Echo on most datasets, suggesting that crop-and-zoom improves the effective resolution and saliency of lesion details and thus provides more reliable attribute evidence. For Calcification, due to severe class imbalance, Acc and Weighted-F1 are less informative, so we do not rely on it as a primary basis for conclusions.

\begin{figure}[t]
    \centering
    \begin{minipage}[t]{0.49\linewidth}
        \centering
        \includegraphics[width=\linewidth]{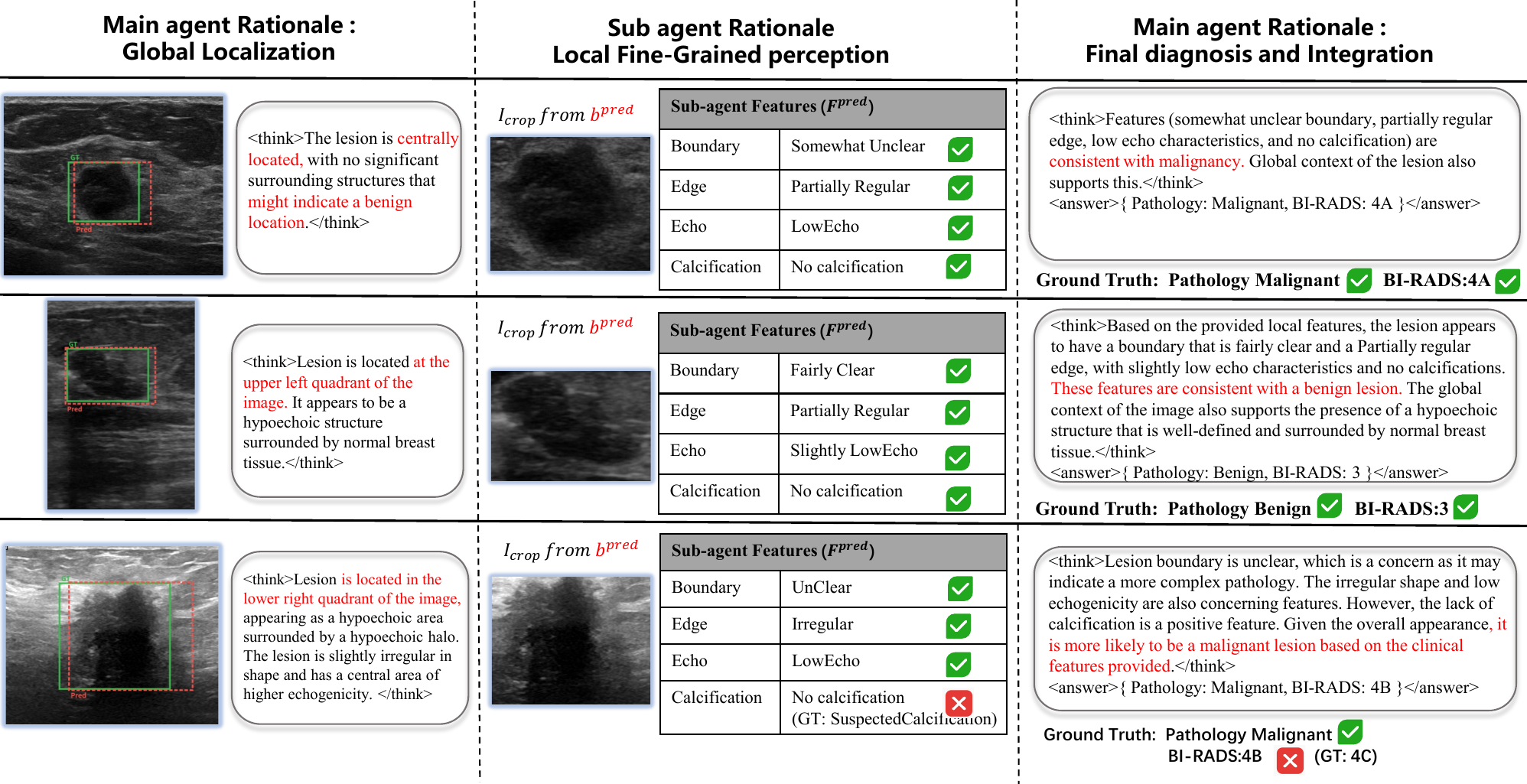}
        \subcaption{Case-wise evidence-chain outputs}
        \label{fig:qual}
    \end{minipage}\hfill
    \begin{minipage}[t]{0.49\linewidth}
        \centering
        \includegraphics[width=\linewidth]{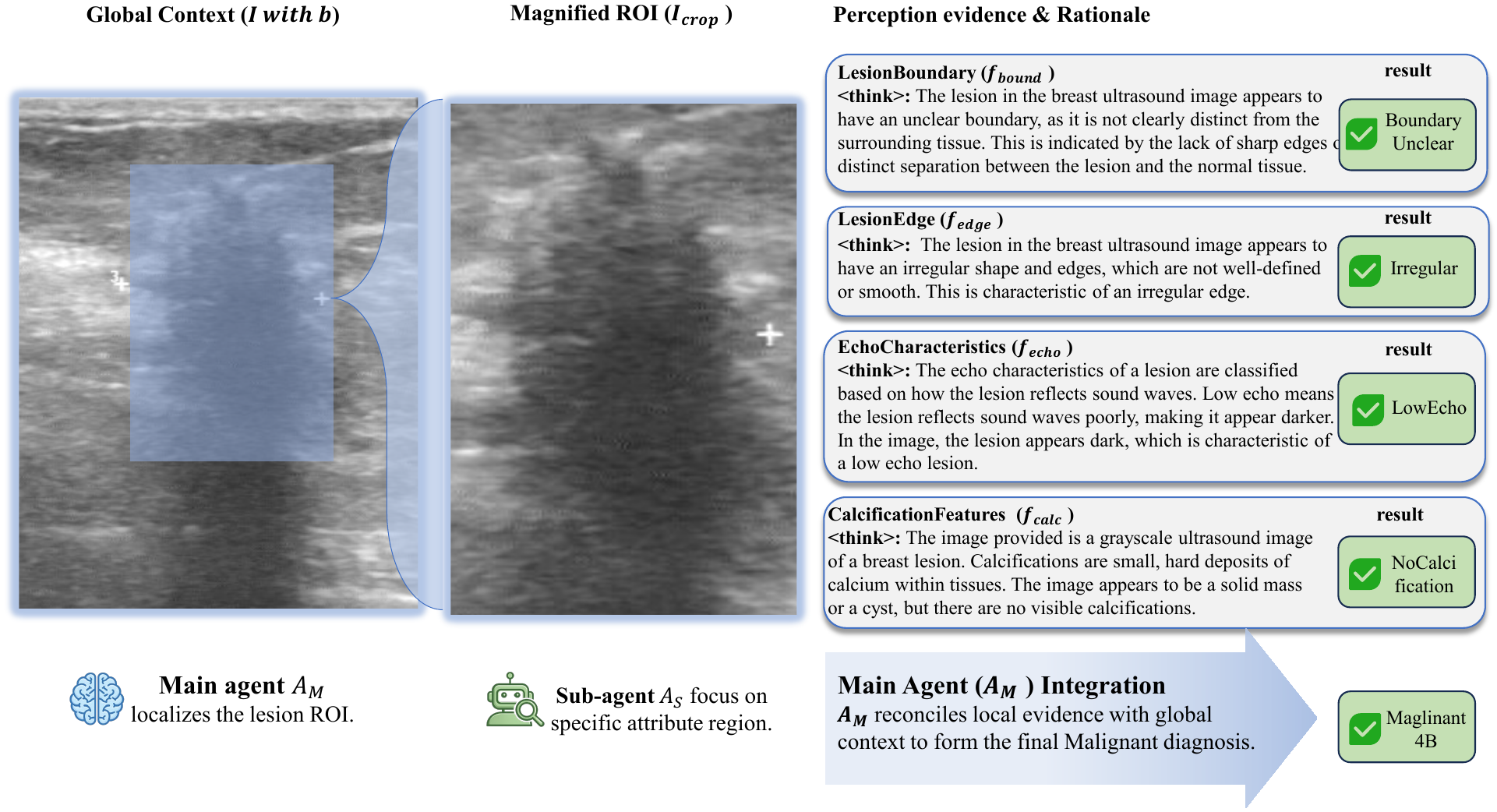}
        \subcaption{Attribute reasoning example}
        \label{fig:attr_qual}
    \end{minipage}
    \caption{Qualitative examples of evidence-chain diagnosis and attribute reasoning traces.}
    \label{fig:qual_all}
\end{figure}

\noindent
\textbf{Qualitative results.}
Fig.~\ref{fig:qual_all} visualizes the ROI$\rightarrow$attribute$\rightarrow$diagnosis pipeline. The main agent proposes an ROI on the full image, the sub-agent predicts structured attributes on the cropped view, and the main agent integrates global context with these attributes to output malignancy and BI-RADS. In challenging cases, attribute evidence helps revise an initial global impression and improves decision consistency. We also observe that isolated noisy attributes do not necessarily dominate the final decision, suggesting that the main agent can reconcile local evidence with global context.

\label{sec:exp:qual}

\section{Conclusion}
\label{sec:conclusion}
We propose a hierarchical multi-agent framework for BUS diagnosis that mirrors clinical reading and forms an explicit ROI$\rightarrow$attribute$\rightarrow$diagnosis evidence-chain. A main agent localizes the lesion and integrates evidence, while a sub-agent recognizes fine-grained attributes on the cropped view. With oracle-guided RL for stable attribute-based reasoning and corrective trajectory self-distillation with spatial supervision for SFT, the resulting end-to-end policy improves diagnostic performance and attribute agreement while producing structured, reviewable intermediate evidence.

Limitations include scarce and imbalanced attribute annotations and sensitivity to noisy intermediate evidence at test time. Future work will expand the attribute taxonomy, incorporate confidence-aware evidence transmission and robust fusion, and evaluate on larger multi-center and multi-view/sequence ultrasound datasets.


\begin{thebibliography}{}
\bibitem{biradsnet}
Zhang, B., Vakanski, A., Xian, M.:
BI-RADS-Net: An explainable multitask learning approach for cancer diagnosis in breast ultrasound images.
In: 2021 IEEE 31st International Workshop on Machine Learning for Signal Processing (MLSP), pp. 1--6 (2021).
\doi{10.1109/MLSP52302.2021.9596314}

\bibitem{buscot}
Yu, H., Li, Y., Niu, Z., et al.:
A chain-of-thought reasoning breast ultrasound dataset covering all histopathology categories.
Scientific Data (Sci Data) (2026).
\doi{10.1038/s41597-026-06702-9}

\bibitem{spak2017birads}
Spak, D.A., Plaxco, J.S., Santiago, L., Dryden, M.J., Dogan, B.E.:
BI-RADS\textsuperscript{\textregistered} fifth edition: A summary of changes.
Diagnostic and Interventional Imaging \textbf{98}(3), 179--190 (2017).
\doi{10.1016/j.diii.2017.01.001}

\bibitem{hayashida2022birads4a}
Hayashida, T., Odani, E., Kikuchi, M., et al.:
Establishment of a deep-learning system to diagnose BI-RADS4a or higher using breast ultrasound for clinical application.
Cancer Science \textbf{113}(10), 3528--3534 (2022).
\doi{10.1111/cas.15511}

\bibitem{carrilero2024describe}
Carrilero-Mardones, M., Parras-Jurado, M., Nogales, A., et al.:
Deep learning for describing breast ultrasound images with BI-RADS terms.
Journal of Imaging Informatics in Medicine \textbf{37}, 2940--2954 (2024).
\doi{10.1007/s10278-024-01155-1}

\bibitem{interpretable}
Cui, K., Liu, W., Wang, D.:
Interpretable diagnosis of breast lesions in ultrasound imaging using deep multi-stage reasoning.
Physics in Medicine and Biology \textbf{69}(21), 215025 (2024).
\doi{10.1088/1361-6560/ad869f}

\bibitem{dualstage}
Bruno, P., Macr\`i, M., Dodaro, C.:
A dual-stage deep learning framework for breast ultrasound image segmentation and classification.
Journal of Medical Systems \textbf{49}, 162 (2025).
\doi{10.1007/s10916-025-02298-6}

\bibitem{medr1}
Lai, Y., Zhong, J., Li, M., et al.:
Med-R1: Reinforcement learning for generalizable medical reasoning in vision-language models.
arXiv:2503.13939 (2025).
\doi{10.48550/arXiv.2503.13939}

\bibitem{medreason}
Li, Y., Tang, F., Li, Y., Zhou, S.K.:
MedReason-R1: Learning to reason for CT diagnosis with reinforcement learning and local zoom.
arXiv:2510.19626 (2025).
\doi{10.48550/arXiv.2510.19626}

\bibitem{medvlmr1}
Pan, J., Liu, C., Wu, J., et al.:
MedVLM-R1: Incentivizing medical reasoning capability of vision-language models (VLMs) via reinforcement learning.
In: Medical Image Computing and Computer Assisted Intervention -- MICCAI 2025.
LNCS, vol. 15966, pp. 337--347. Springer, Cham (2026).
\doi{10.1007/978-3-032-04981-0_32}

\bibitem{deepeyes}
Zheng, Z., Yang, M., Hong, J., et al.:
DeepEyes: Incentivizing ``thinking with images'' via reinforcement learning.
arXiv:2505.14362 (2025).
\doi{10.48550/arXiv.2505.14362}

\bibitem{vitar}
Chen, K., Rui, S., Jiang, Y., et al.:
Think twice to see more: Iterative visual reasoning in medical VLMs.
arXiv:2510.10052 (2025).
\doi{10.48550/arXiv.2510.10052}

\bibitem{pateria2021hrlsurvey}
Pateria, S., Subagdja, B., Tan, A.-H., Quek, C.:
Hierarchical reinforcement learning: A comprehensive survey.
ACM Computing Surveys \textbf{54}(5), Article 109 (2021).
\doi{10.1145/3453160}

\bibitem{medagent}
Wang, Z., Wu, J., Cai, L., et al.:
MedAgent-Pro: Towards evidence-based multi-modal medical diagnosis via reasoning agentic workflow.
arXiv:2503.18968 (2025).
\doi{10.48550/arXiv.2503.18968}

\bibitem{treereasoning}
Peng, Q., Cui, J., Xie, J., Cai, Y., Li, Q.:
Tree-of-Reasoning: Towards complex medical diagnosis via multi-agent reasoning with evidence tree.
In: Proceedings of the 33rd ACM International Conference on Multimedia (MM '25), pp. 1744--1753 (2025).
\doi{10.1145/3746027.3755418}

\bibitem{busi}
Al-Dhabyani, W., Gomaa, M., Khaled, H., Fahmy, A.:
Dataset of breast ultrasound images.
Data in Brief \textbf{28}, 104863 (2020).
\doi{10.1016/j.dib.2019.104863}


\bibitem{busbra}
G\'omez-Flores, W., et al.:
BUS-BRA: A breast ultrasound dataset for assessing computer-aided diagnosis systems.
Medical Physics \textbf{51}(4), 3110--3123 (2024).
\doi{10.1002/mp.16812}

\bibitem{budiat}
Yap, M.H., Pons, G., Marti, J., Ganau, S., Sentis, M., Zwiggelaar, R., Davison, A.K., Marti, R.:
Automated breast ultrasound lesions detection using convolutional neural networks.
IEEE Journal of Biomedical and Health Informatics \textbf{22}(4), 1218--1226 (2018).
\doi{10.1109/JBHI.2017.2731873}

\bibitem{breast}
Paw{\l}owska, A., {\'C}wierz-Pie{\'n}kowska, A., Domalik, A., Jagu{\'s}, D., et al.:
Curated benchmark dataset for ultrasound based breast lesion analysis.
Scientific Data \textbf{11} (2024).
\doi{10.1038/s41597-024-02984-z}

\bibitem{shao2024deepseekmath}
Shao, Z., Wang, P., Zhu, Q., et al.:
DeepSeekMath: Pushing the limits of mathematical reasoning in open language models.
arXiv preprint arXiv:2402.03300 (2024).
\doi{10.48550/arXiv.2402.03300}

\bibitem{kappa}
Cohen, J.:
A coefficient of agreement for nominal scales.
Educational and Psychological Measurement \textbf{20}(1), 37--46 (1960).
\doi{10.1177/001316446002000104}

\end{thebibliography}
\end{document}